\documentclass[11pt]{article}
\usepackage[margin=1in]{geometry}
\usepackage[T1]{fontenc}
\usepackage{newtxtext,newtxmath}
\usepackage{graphicx}
\usepackage[section]{placeins}
\usepackage{booktabs}

\usepackage{amsmath,amssymb}
\usepackage{xcolor}
\usepackage{natbib}
\usepackage{hyperref}
\hypersetup{colorlinks=true,linkcolor=blue,citecolor=blue,urlcolor=blue}
\usepackage{microtype}
\newcommand{\gold}{\textsc{gold}}
\newcommand{\cons}{\textsc{consensus}}
\newcommand{\review}{\textsc{review}}
\newcommand{\vlmstar}{VLM$^{\star}$}

\newcommand{\zenododoi}{10.5281/zenodo.21753106}
\newif\ifdepositlive \depositlivetrue

\newcommand{\repourl}{https://github.com/Solid-Energy-Systems/verdict}
\newif\ifrepolive \repolivetrue
\title{VERDICT: Agreement Beats Pixel-Space Verification in Real-Document OCSR}
\author{Yani Guan\thanks{These authors contributed equally to this work.}, Dengpan Dong\footnotemark[1], Shuang Luo\footnotemark[1], Zi Wei, Joah Han,
\\ Dan Hannah, Yumin Zhang\thanks{Corresponding author: \texttt{yumin.zhang@ses.ai}, \texttt{qichao.hu@ses.ai}, \texttt{kang.xu@ses.ai}}, Qichao Hu\footnotemark[2], Kang Xu\footnotemark[2]\\
SES AI Corporation\\
}
\date{}

\begin{document}
\maketitle

\begin{abstract}
\noindent Optical Chemical Structure Recognition (OCSR), which converts 2D molecular depictions in the published literature into SMILES, is increasingly important for constructing large-scale chemical training datasets. However, automation at that scale
requires identifying which predictions are unreliable, a decision that must be
made without ground truth.  Three label-free reliability signals were compared: model confidence, re-rendering similarity and agreement among different recognizers. We evaluated $263$ molecular depictions from ACS journals for which definitive ground truth was available. Pixel-space re-rendering performed little better than chance (AUROC
$0.547$, $95\%$ CI $[0.465,0.629]$), and an oracle-tuned threshold on that signal
reduced correct labels per input image from $0.745$ to $0.205$. Agreement among
four architecturally distinct recognizers instead reached an AUROC of $0.916$
($[0.880,0.952]$). The two-of-four rule accepted $81.7\%$ of the images at $88.8\%$ precision, while the three-of-four rule accepted $52.1\%$ at $98.5\%$ precision. The same pattern held on CLEF-IP, UOB, and USPTO. This distinction is obscured on synthetic benchmarks, where re-rendered predictions naturally resemble their inputs. A chemical filter removed $2{,}193$ false agreements on wildcards and R-group fragments. After filtering, the three-of-four rule rejected all $68$ generic depictions. VERDICT was then applied to PMC Open Access and produced $6{,}146$ structure labels for $4{,}833$ molecules. Chemist adjudication of $400$ released labels in two independent samples yielded precisions of $0.995$ for the three-of-four tier and $0.958$ for the two-of-four tier. VERDICT therefore enables the generation of validated labels for multimodal molecular databases linking structure images, machine-readable molecular representations, and information from source publications. In SES AI's Molecular Universe platform, VERDICT further serves as a reliable image-based interface for searching and retrieving molecular records.

\end{abstract}

\section{Introduction}
A substantial fraction of chemical multi-modal knowledge remains encoded in molecular structure figures within papers and patents, placing it out of reach of chemical model training. Optical Chemical Structure Recognition (OCSR) offers a way to convert 2D molecular depictions into machine-readable structures i.e. SMILES, but no ground truth exists for a figure taken from an arbitrary paper. Therefore, a reliable system around those models is required to decide on its own which of its predictions to keep to build a high quality multi-modal molecular resources. 

At the model level, specialized recognizers continue to be developed, including DECIMER \citep{decimer,decimer2}, MolScribe \citep{molscribe}, and MolNexTR \citep{molnextr}, all of which report exact-match accuracies above $0.9$ on synthetic benchmarks. Recently, a rapidly growing line of chemistry-specific and general-purpose vision language models (VLMs) has followed, including reasoning-trace approaches that emit an explicit graph-traversal derivation before the final string \citep{chemvlm,molparser,gtrcot,markushglyph}. Real documents, however, present depictions of unpredictable style and size, embedded in dense page context from which they must first be isolated, so the resulting crops are frequently far harder to read than benchmark images. The effect was measured in a companion study \citep{ft}, where a fine-tuned VLM recognizer achieved an average exact-match accuracy of $0.940$ across four rendered conditions but only $0.231$ across four real-document sets. Reliability is therefore a requirement distinct from recognizer accuracy, and one that has to be met at the system level. Existing system-level work has addressed integration rather than validation: document-level pipelines assemble chemical records from molecular figures and surrounding text \citep{chemdataextractor,rxnscribe,openchemie}, which raises the cost of an incorrect structure without introducing any criterion for rejecting one. Among the recognizers themselves, MolScribe supplies a per-prediction confidence score \citep{molscribe}, but it is single-engine and has no counterpart in the others. For a reliable system, three ground-truth-free signals are available in principle: model confidence \citep{hendrycks,guo2017calibration}, self-verification through re-rendering, and agreement among independent recognizers \citep{dietterich,selfconsistency,snorkel}. Each carries a plausible failure mode. Confidence scores are unavailable or incomparable across many OCSR engines; re-rendering similarity may reflect depiction style rather than chemical correctness; and cross-model agreement can preserve errors shared among recognizers. These signals have not been systematically evaluated under a common protocol on real OCSR images, and existing real-image benchmarks remain limited in scale or dominated by patent-derived depictions \citep{clef,uob,staker}. Consequently, the field lacks both a validated criterion for selective acceptance and a reliable strategy for constructing large-scale, journal-domain molecular corpora.
 
In this work, we present a reliable system designed to recognize when a difficult crop has defeated the recognizers. Its consensus mechanism accounts for variability in real depictions, including style, scale, and how cleanly a structure separates from the surrounding page. On $263$ ACS depictions with verified ground truth, pixel-space round-trip verification is close to uninformative (AUROC $0.547$, $95\%$ CI $[0.465,0.629]$), whereas agreement among four architecturally distinct recognizers is strongly discriminative ($0.916$, $[0.880,0.952]$). The same ordering is reproduced on CLEF-IP, UOB, and USPTO. Pixel comparison measures depiction style, while agreement is assessed on molecular identity and is largely indifferent to it. VERDICT reconciles engine outputs using molecular identity rather than exact SMILES-string matching. A substance filter then removes false agreements in which multiple engines converge on wildcard- or R-group-containing fragments rather than complete molecules. It accepts $81.7\%$ of images at $0.888$ precision and $52.1\%$ at $0.985$, in a median $3.5$\,s per image. Therefore, VERDICT establishes which crops were read correctly and abstains on the rest. As an application, we ran this framework over PubMed Central (PMC) Open Access, where it yielded $6{,}146$ structure labels for $4{,}833$ distinct molecules, with chemist evaluation of $400$ labels drawn in two independent samples demonstrating a precision of at least $0.958$. This supports the construction of gold-standard datasets of real labeled depictions, helping to close the synthetic-to-real gap \citep{ft}. A
validated gate can therefore produce such datasets from unlabeled literature, allowing the corpus generated by current recognizers to train future OCSR models. Besides, when incorporated into a knowledge base, the same labels link each structure to the publication, properties, and provenance recorded alongside it. Consequently, a molecule depicted in one paper can be connected to relevant information reported elsewhere. 

\FloatBarrier

%% 2. 
\section{VERDICT System Framework}\label{sec:verdict}

%% 2.1
\subsection{Confidence of VERDICT gate}
\label{par:contamination}
\begin{figure}[t]
\centering
\includegraphics[width=\textwidth]{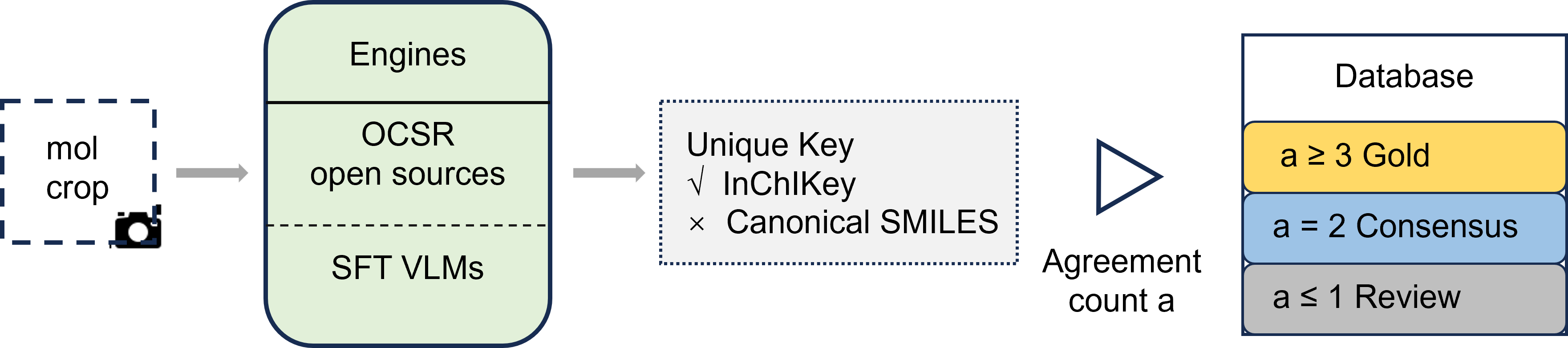}
\caption{VERDICT groups four recognizer outputs by molecular identity and
assigns each image to an agreement tier.}
\label{fig:system}
\end{figure}

The reliability of the VERDICT framework rests on three safeguards. First, predictions are compared within the
consensus gate using stereochemistry-preserving InChIKeys rather than SMILES strings, because SMILES is a serialization rather than an identifier; a single molecule may therefore have more than one valid SMILES representation. VERDICT therefore reconciles predictions at the level of molecular identity. More specifically, four recognizers—DECIMER, MolScribe, MolNexTR, and the supervised fine-tuned \vlmstar{} \citep{ft}—each produce a prediction. The predicted SMILES strings are then canonicalized and converted to stereochemistry-preserving InChIKeys for comparison. Accepted images are assigned to tiers based on agreement count: \gold{} for $a\geq3$ and \cons{} for $a=2$, with $a\leq1$ routed to \review{} and never emitted as a label (Figure~\ref{fig:system}).

Beyond molecular identity matching, the reliability of the consensus gate also depends on whether the recognizers exhibit distinct rather than strongly correlated error patterns. Table~\ref{tab:enginesolo}, reproduced from our companion paper \citep{ft}, shows that four different recognizers, which differ in architecture, tokenization, and training data, exhibit different strengths under different conditions. To provide diverse perspectives, error independence is more important than apparent complementarity in aggregate performance. However, rank reversals across conditions cannot establish that the errors are uncorrelated. This question is not settled by construction and is therefore measured directly in \S\ref{sec:signals}.\footnote{Scores in Table~\ref{tab:enginesolo} are taken from the companion study \citep{ft} and use canonical-string exact match on all $331$ ACS images, whereas the gate evaluation uses identity-key match on the $263$ images with verified ground truth. The two are not directly comparable. The common adapter interface is described in Appendix~\ref{app:engine}.}

Finally, we measured molecule-level overlap of the ACS evaluation set and PMC corpus with the training data
used for the in-house voter. None of the $257$ distinct ACS molecules appeared in the training dataset, whereas $15.0\%$ of the molecules in the PMC corpus overlapped with the training pool. Accordingly, the released records carry both the supporting engines and a training-overlap flag, and restricting acceptance to the three public recognizers entirely removes dependence on the in-house voter.

\begin{table}[t]
\centering
\small
\caption{The four recognizers perform differently across rendered and real-document datasets.}
\label{tab:enginesolo}
\setlength{\tabcolsep}{5pt}
\begin{tabular}{lcccccc}
\toprule
& \multicolumn{2}{c}{rendered}
& \multicolumn{4}{c}{real documents} \\
\cmidrule(lr){2-3}
\cmidrule(lr){4-7}
recognizer & clean & degraded & ACS & CLEF-IP & UOB & USPTO \\
\midrule
DECIMER
& $.958$ & $.550$ & $.296$ & $.732$ & $.866$ & $.586$ \\
MolScribe
& $.716$ & $.272$ & $.583$ & $.726$ & $.824$ & $.766$ \\
MolNexTR
& $.802$ & $.340$ & $.520$ & $.732$ & $.828$ & $.796$ \\
\vlmstar{}
& $.958$ & $.956$ & $.459$ & $.610$ & $.830$ & $.732$ \\
\bottomrule
\end{tabular}
\end{table}

\FloatBarrier

%% 2.2
\subsection{Mechanism of VERDICT consensus}\label{sec:keybug}
Here, we describe the first component of the confidence gate in greater detail: representing molecular identity using InChIKey. Let engine $e$ emit the raw string $\hat y_e$, and define
\[
\kappa(\hat y)=
\begin{cases}
\mathrm{InChIKey}(\hat y), & \text{RDKit parses } \hat y \text{ and InChI generation succeeds},\\
\bot, & \text{otherwise.}
\end{cases}
\]
Predictions mapped to $\bot$ are discarded, and the winning key $\kappa^{\star}$ is the mode of the remaining keys, with agreement count $a=|\{e:\kappa(\hat y_e)=\kappa^{\star}\}|$. Voting on molecular identity in this way accounts for the representational variability among equivalent SMILES strings.

In addition, VERDICT accepts a key only when a valid InChIKey is generated, no dummy atom is present, and the structure contains at least six heavy atoms. These criteria exclude incomplete or generic structures, such as wildcard-containing structures and R-group fragments. In a parallel ablation experiment, $\kappa$ was relaxed to fall back on canonical SMILES when InChI generation failed. This relaxation increased the number of accepted labels in the PMC corpus from $6{,}881$ to $9{,}074$, an apparent gain of $32\%$. However, all $2{,}193$ additional outputs were wildcard-containing structures or R-group fragments, and none represented a complete molecule. Table~\ref{tab:thresh} shows that the corpus size is insensitive to the exact heavy-atom threshold, varying by only a few percent across neighboring values. This filter therefore also defines the boundary of the task. Among $68$ ACS depictions whose ground truth is a generic structure, one image reached $a=2$, and none reached $a\geq3$. Thus, the high-trust tier admitted no out-of-scope structures on this set, although the sample is too small to characterize performance on dense patent Markush drawings.

\begin{table}[t]\centering\small
\caption{Sensitivity to the heavy-atom threshold in the substance check, applied to
the $6{,}881$ strict-InChIKey labels. The deployed value is $\geq6$.}
\label{tab:thresh}
\begin{tabular}{lcccccccc}
\toprule
min.\ heavy atoms & $\geq1$ & $\geq2$ & $\geq3$ & $\geq4$ & $\geq5$ & $\geq6$ & $\geq7$ & $\geq8$\\
\midrule
labels & $6{,}881$ & $6{,}745$ & $6{,}701$ & $6{,}660$ & $6{,}545$ & $6{,}320$ & $6{,}178$ & $5{,}992$\\
\gold{} slice & $3{,}215$ & $3{,}213$ & $3{,}212$ & $3{,}207$ & $3{,}166$ & $3{,}097$ & $3{,}020$ & $2{,}917$\\
\bottomrule
\end{tabular}
\end{table}

\FloatBarrier

%% ===========================================================================
%% SECTION 3
%%

%%   3.1  the asymmetry itself, and why it orders all four signals
%%   3.2  a prediction it forces: agreement cannot raise accuracy, only
%%        partition. Confirmed. Previously presented as a separate surprise.
%%   3.3  a prediction it forces: style-independence implies cross-domain
%%        transfer of thresholds. Confirmed. Previously "more datasets".
%%   3.4  a refinement it forces: the carrier is the agreement FRACTION, not
%%        the count. The fifth-engine experiment tests this rather than being
%%        an ablation, and the awkward ACS precision drop becomes the evidence.
%%   3.5  the operating points, as the application of the above.
%%
%% ===========================================================================

%% 3.
\section{VERDICT Performance and Discussions}\label{sec:signals}
Here, $263$ of $331$ ACS journal depictions that have ground-truth structures are used to compare three families of ground-truth-free reliability methods: intrinsic confidence, self-verification, and multi-engine agreement. Self-verification is evaluated in two forms—pixel-space round-trip similarity and identity-level render-then-recognize consistency—yielding four operational signals in total. These signals differ in the representation spaces in which they are computed. Results and discussion below show that agreement is valuable not because it corrects predictions, but because identity-level, style-insensitive agreement partitions existing predictions into reliability tiers with transferable coverage–precision operating points.

%% ---------------------------------------------------------------------------
\subsection{Comparison of four ground-truth-free signals}
\label{sec:signalcompare}

\begin{figure}[t]
\centering
\includegraphics[width=\textwidth]{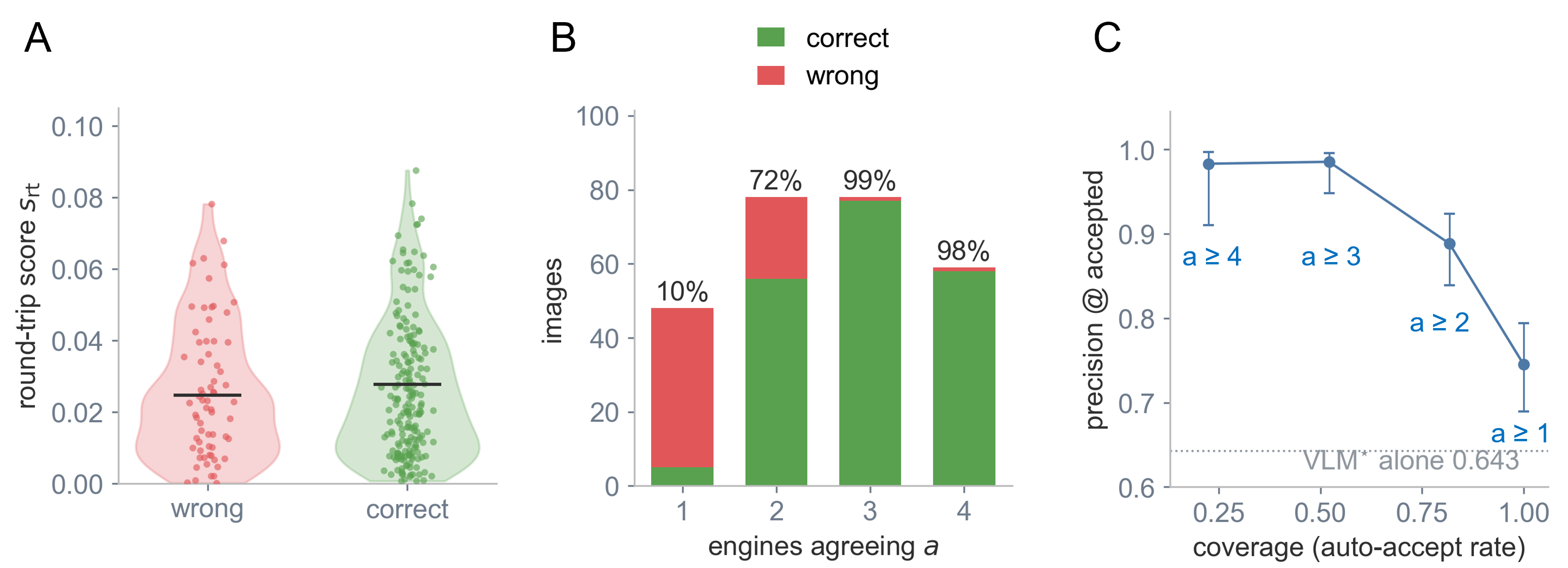}
\caption{Agreement separates correct from incorrect predictions, while pixel-space re-rendering does not. Results are shown for the $263$ ACS depictions with concrete ground truth.}
\label{fig:signals}
\end{figure}

The first signal is pixel-space round-trip verification, which re-renders a candidate structure and compares the resulting image with the source crop. It reaches an AUROC of $0.547$ ($95\%$ CI $[0.465,0.629$]), with the interval including chance performance. Correct and incorrect predictions occupy nearly the same score range, as shown in the left panel of Figure~\ref{fig:signals}. Even an oracle-tuned threshold selected using the ground truth yields only $0.205$ correct labels per input image, compared with $0.745$ when all predictions are emitted. This failure arises from the space in which the comparison is made. Depictions of the same molecule can differ in bond style, spacing, font, and layout. For example, a clean RDKit rendering may differ substantially from a crop extracted from a journal figure, even when both represent the same molecule and are readily recognized as such by a human observer. Pixel-space similarity is therefore dominated by depiction style rather than chemical correctness. 

On the other hand, multi-engine agreement instead compares the molecular identity recovered by each recognizer. Each prediction is reduced to an InChIKey, removing representational differences among equivalent SMILES strings, although not the recognition errors of the engines themselves. The right panel of Figure~\ref{fig:signals} shows a clear separation between correct and incorrect predictions. Agreement count reaches an AUROC of $0.916$ ($95\%$ CI $[0.880,0.952]$), and accuracy rises from approximately $10\%$ among predictions supported by a single engine to $98$--$99\%$ among those supported by at least three engines.

The other two signals—intrinsic confidence and render-then-recognize self-consistency—fall between pixel-space round-trip verification and multi-engine agreement. MolScribe's intrinsic confidence reaches an AUROC of $0.666$. Render-then-recognize self-consistency ranges from $0.566$ to $0.808$ across engines, with the per-engine values reported in Table~\ref{tab:selfconsistency}. Render-then-recognize compares the identity of the original prediction with that recovered from its re-rendering, whereas intrinsic confidence is computed from the model's own predictive distribution. Neither relies on direct pixel similarity, which is consistent with their higher AUROC point estimates relative to pixel-space round-trip verification. However, neither combines evidence from multiple recognizers, consistent with their lower AUROC values than multi-engine agreement.

The ordering in Table~\ref{tab:signals} can therefore be understood in terms of two properties rather than as a comparison among unrelated heuristics: whether the signal relies directly on depiction-level pixel similarity and whether it combines evidence from more than one recognizer. Table~\ref{tab:selfconsistency} further shows that self-consistency is not equivalent to informativeness. DECIMER reproduces its own molecular identity on $92.2\%$ of images but provides little information about correctness, whereas MolNexTR is self-consistent on $70.1\%$ and provides the strongest self-consistency signal among the evaluated engines. Reproducing a previous output demonstrates prediction stability, not necessarily correctness.

\begin{table}[t]
\centering
\small
\caption{Signal strength follows the space of evaluation. The two signals computed on molecular identity exceed the one computed on pixels, and the one signal drawing on multiple models exceeds those drawing on one. AUROC intervals are $95\%$ DeLong intervals and $p$-values are two-sided Mann--Whitney tests \citep{delong}. The render-then-recognize row reports the range across the three engines of Table~\ref{tab:selfconsistency}.}
\label{tab:signals}
\setlength{\tabcolsep}{4pt}
\begin{tabular}{lccccc}
\toprule
signal
& mean, correct
& mean, wrong
& range
& AUROC
& $p$ \\
\midrule
round-trip $s_{\mathrm{rt}}$
& $0.0277$
& $0.0247$
& $0.0005$--$0.0982$
& $0.547$ {\small$[.465,.629]$}
& $0.25$ \\
MolScribe confidence
& $0.876$
& $0.810$
& $0.001$--$0.927$
& $0.666$ {\small$[.598,.734]$}
& $2\times10^{-5}$ \\
render-then-recognize
& ---
& ---
& $\{0,1\}$
& $0.566$--$0.808$
& --- \\
agreement count $a$
& ---
& ---
& $1$--$4$
& $0.916$ {\small$[.880,.952]$}
& $<\!10^{-20}$ \\
\bottomrule
\end{tabular}
\end{table}

\begin{table}[t]
\centering
\small
\caption{Per-engine render-then-recognize self-consistency. The engine that most often reproduces its own output is the least informative one, so self-consistency measures stability rather than correctness.}
\label{tab:selfconsistency}
\begin{tabular}{lccc}
\toprule
signal & self-consistent & AUROC ($95\%$) &
precision @ coverage \\
\midrule
DECIMER
    & $236/256=0.922$
    & $0.566$ {\small$[.534,.597]$}
    & $0.411$ @ $89.7\%$ \\
MolScribe
    & $219/254=0.862$
    & $0.721$ {\small$[.664,.779]$}
    & $0.817$ @ $83.3\%$ \\
MolNexTR
    & $178/254=0.701$
    & $0.808$ {\small$[.758,.856]$}
    & $0.831$ @ $67.7\%$ \\
\midrule
agreement $a\!\geq\!2$
    & ---
    & $0.916$ {\small$[.880,.952]$}
    & $0.888$ @ $81.7\%$ \\
\bottomrule
\end{tabular}
\end{table}

%% ---------------------------------------------------------------------------
\subsection{Abstention plays a role in agreements among engines}
\label{sec:partition}

The preceding analysis highlights a limitation. Agreement indicates whether a reading is reliable but it does not produce a new prediction. When required to provide an output for every image, a quorum can only select among structures that its members have already predicted. Consequently, its accuracy should remain close to that of its best-performing member. Table~\ref{tab:single-engine} confirms this on ACS. The four-engine quorum reaches $0.745$ exact match against $0.707$ for MolScribe, the strongest individual recognizer, a gain of less than four points. The four columns of Table~\ref{tab:crossset} labelled "quorum all" and "best engine" confirm it across four real-document sets, where the difference ranges from $-0.6$ to $+3.4$ percentage points. On USPTO the quorum is slightly worse than the best single engine.

The same account predicts the opposite behavior once abstention is allowed, because the quantity the signal carries is then permitted to act. Requiring at least three agreeing engines reduces the error rate among accepted predictions by a factor of $1.8$ on CLEF-IP, $3.0$ on UOB, $7.7$ on USPTO, and $11.8$ on ACS, while retaining between $52.5\%$ and $95.0\%$ of the images. Accuracy moves by at most $3.4$ points. Errors among accepted predictions fall by up to an order of magnitude.

Reconciliation is therefore a partition rather than a correction. It sorts predictions into a reliable subset and an unreliable one, and it does not repair the second. The distinction has a practical consequence. A quorum evaluated with abstention disabled will appear to add almost nothing, and a quorum reported as an accuracy improvement will be credited with something it does not do.

\begin{table}[t]
\centering
\small
\caption{With abstention disabled, the four-engine quorum exceeds the best individual recognizer on ACS by less than four points. The dash for \vlmstar{} records that it emits a string for every image, so its answer rate is not defined against the same denominator as the parsing-limited engines.}
\label{tab:single-engine}
\begin{tabular}{lccc}
\toprule
recognizer & exact match & $95\%$ Wilson CI & answer rate \\
\midrule
DECIMER
    & $0.376$
    & $[.320,.436]$
    & $97.3\%$ \\
MolNexTR
    & $0.616$
    & $[.556,.673]$
    & $79.5\%$ \\
\vlmstar{}
    & $0.643$
    & $[.583,.698]$
    & --- \\
MolScribe
    & $0.707$
    & $[.650,.759]$
    & $94.7\%$ \\
\midrule
four-engine quorum
    & $0.745$
    & $[.689,.794]$
    & $100\%$ \\
\bottomrule
\end{tabular}
\end{table}

%% ---------------------------------------------------------------------------
\subsection{From ACS to other depiction domains}
\label{sec:crossset}

If the signal is insensitive to how a molecule is drawn, then thresholds calibrated on one depiction domain should hold on others. The prediction is testable and consequential. Thresholds were calibrated on ACS journal figures, whereas the corpus of \S\ref{sec:corpus} is built from PMC articles, so the operating points are applied outside the domain that produced them.

The gate was applied without modification to CLEF-IP patent figures, UOB hand-drawn structures, and a USPTO-derived set, using one implementation and the same identity-key scoring rule. Table~\ref{tab:crossset} reports the result. At $a\!\geq\!3$, precision lies between $0.958$ and $0.992$ on all four sets. The two domains furthest from the calibration set are not the weak cases: UOB, which is hand-drawn, reaches $0.992$, and USPTO, which is patent-derived, reaches $0.989$, both above the $0.978$ obtained on ACS itself.

This transfer is expected under the explanation in \S\ref{sec:signalcompare} and would be difficult to explain if the signal depended on depiction style. The cross-set experiment therefore tests the proposed mechanism rather than merely extending dataset coverage.

Two comparability notes apply to Table~\ref{tab:crossset}. The unified implementation used here reproduces the archived ACS results to within $0.8$ percentage points on both axes, the largest single differences being $0.76$ points of coverage at $a\!\geq\!2$ and $0.71$ points of precision at $a\!\geq\!3$. This accounts for the $0.985$ reported for the \gold{} tier in Table~\ref{tab:gate} against $0.978$ here. The non-ACS runs also use the first $500$ archived rows of each benchmark rather than random samples, so the values in Table~\ref{tab:crossset} are not sampling estimates for the full benchmarks, as discussed in \S\ref{sec:limits}.

\begin{table}[t]
\centering
\small
\caption{Thresholds calibrated on ACS transfer to patent and hand-drawn depictions. Raw accuracy moves by at most $3.4$ points, while the error rate among accepted predictions falls by a factor of $1.8$ to $11.8$ once abstention at $a\!\geq\!3$ is allowed.}
\label{tab:crossset}
\setlength{\tabcolsep}{3.5pt}
\begin{tabular}{lrccccc}
\toprule
set & $n$ & quorum all & best engine & difference
& $a\!\geq\!2$ cov./prec.
& $a\!\geq\!3$ cov./prec. \\
\midrule
ACS
& $263$ & $0.741$ & $0.707$ & $+3.4$
& $0.810$ / $0.892$
& $0.525$ / $0.978$ \\
CLEF-IP
& $444$ & $0.923$ & $0.917$ & $+0.7$
& $0.982$ / $0.933$
& $0.854$ / $0.958$ \\
UOB
& $500$ & $0.976$ & $0.960$ & $+1.6$
& $0.994$ / $0.978$
& $0.950$ / $0.992$ \\
USPTO
& $494$ & $0.915$ & $0.921$ & $-0.6$
& $0.923$ / $0.974$
& $0.763$ / $0.989$ \\
\bottomrule
\end{tabular}
\end{table}

\FloatBarrier

%% ---------------------------------------------------------------------------
\subsection{Agreement fraction is the operating specification}
\label{sec:quorumsize}

Agreement counts among voters with abstention produce good predictions on ACS depictions as well as CLEF-IP, UOB and USPTO. However, the count is a convenient label rather than the quantity that carries the information. Three of four requires $75\%$ agreement. Three of five requires $60\%$. Rows sharing a
count label therefore describe different acceptance rules once the roster changes, and the effect is measurable.

OCSRGlyph \citep{markushglyph}, developed for Markush and patent-style depictions and architecturally distinct from the other four recognizers, was added as a fifth voter with no other component changed.
Table~\ref{tab:fifth-voter} reports both rosters at $a\!\geq\!2$, $a\!\geq\!3$ and $a\!\geq\!4$. Read by count label, the fifth engine appears to behave inconsistently. On USPTO, coverage at $a\!\geq\!3$ rises from $76.3\%$ to $90.5\%$ with precision unchanged at $0.989$. On CLEF-IP it rises from $85.4\%$ to $94.8\%$ with precision rising from $0.958$ to $0.962$. On ACS coverage also rises, from $52.5\%$ to $67.3\%$, but precision falls from $0.978$ to $0.955$. The fall amounts to three errors in $138$ accepted images against eight in $177$, with overlapping Wilson intervals $[.938,.993]$ and $[.913,.977]$, so it is not statistically distinguishable in any case. It is also not a like-for-like comparison, since $a\!\geq\!3$ of five is a weaker requirement than $a\!\geq\!3$ of four.

When interpreted as operating curves, the results are consistent. Table~\ref{tab:fifth-voter} shows that the five-engine roster supplies points that dominate the four-engine curve at its high-precision end. On ACS it reaches $0.991$ precision at $42.2\%$ coverage, against $0.983$ at $22.8\%$ for four-engine unanimity: higher precision and nearly double the coverage. On CLEF-IP it reaches $0.984$ at $81.8\%$ against $0.979$ at $53.1\%$. On USPTO the fifth engine trades rather than dominates, exchanging $1.000$ precision at $43.1\%$ coverage for $0.989$ at $75.5\%$.

The operating specification of the gate is therefore an agreement fraction together with a roster size, not a vote count. A threshold must be recalibrated whenever the roster changes, and a gate that reports a count without the pool size has not specified what it accepts.

\begin{table}[t]
\centering
\small
\caption{Adding a fifth recognizer extends the coverage--precision curve rather than shifting it. Read by count label the fifth engine appears to reduce ACS precision, because $a\!\geq\!3$ of five requires only $60\%$ agreement where $a\!\geq\!3$ of four requires $75\%$. Read as a curve, the five-engine roster dominates at the high-precision end on ACS and CLEF-IP. UOB is omitted because archived five engine predictions are unavailable for that set.}
\label{tab:fifth-voter}
\setlength{\tabcolsep}{4pt}
\begin{tabular}{llccc}
\toprule
set & quorum
& $a\!\geq\!2$ cov./prec.
& $a\!\geq\!3$ cov./prec.
& $a\!\geq\!4$ cov./prec. \\
\midrule
USPTO & four engines
& $0.923$ / $0.974$
& $0.763$ / $0.989$
& $0.431$ / $1.000$ \\
USPTO & five engines
& $0.968$ / $0.977$
& $0.905$ / $0.989$
& $0.755$ / $0.989$ \\
\midrule
CLEF-IP & four engines
& $0.982$ / $0.933$
& $0.854$ / $0.958$
& $0.531$ / $0.979$ \\
CLEF-IP & five engines
& $0.998$ / $0.953$
& $0.948$ / $0.962$
& $0.818$ / $0.984$ \\
\midrule
ACS & four engines
& $0.810$ / $0.892$
& $0.525$ / $0.978$
& $0.228$ / $0.983$ \\
ACS & five engines
& $0.837$ / $0.896$
& $0.673$ / $0.955$
& $0.422$ / $0.991$ \\
\bottomrule
\end{tabular}
\end{table}

\FloatBarrier

%% ---------------------------------------------------------------------------
\subsection{Operating points}
\label{sec:operating}

Agreement fraction among voters with abstention demonstrates the ability to generate high quality labeled depictions as shown in Table~\ref{tab:gate}. Two thresholds are retained. First, the two-of-four rule accepts $81.7\%$ of images at $0.888$ precision and is used for corpus construction, where reach determines how much of the literature is recovered. Second, the three of-four rule accepts $52.1\%$ at $0.985$ and is used where an incorrect structure costs more than a missing one. Unanimity is not a third option: coverage falls to $22.4\%$ while observed precision does not rise, which is consistent with the four-value resolution noted in \S\ref{sec:signalcompare}. The first row of Table~\ref{tab:gate} gives the reference point. A single recognizer answering every image reaches $0.643$, so the gate converts a recognizer that is wrong on one image in three into a labeling process that is wrong on one in sixty-seven at half coverage.

Both thresholds were selected on the images used to report them, so observed precision overstates what a fresh sample would yield. The final column of Table~\ref{tab:gate} gives simultaneous distribution-free lower bounds, obtained from Clopper--Pearson intervals with a finite-family Learn-then-Test correction \citep{clopperpearson,learnthentest}. These are $0.829$ for \cons{} and $0.940$ for \gold{}. The bounds, rather than the point estimates, are what the gate can be held to.

Computational cost is determined by the recognizers. The four engines run in parallel on one GPU node at a median of $3.49$\,s per image and a $p95$ of $9.27$\,s, or about one GPU-hour per thousand images on a single H200. Both tiers are read from the same engine outputs, so tightening the threshold adds no computation. References to zero marginal cost in this paper refer to API charges and rather than computational cost. Only the knowledge-base extraction of \S\ref{sec:kb} requires a paid model call.

What the gate does not do is recover the images it rejects. At $a\!\geq\!2$ these number $48$ of $263$. The final row of Table~\ref{tab:gate} previews what a frontier model recovers from them, which \S\ref{sec:backstop} examines in full.

\begin{table}[t]
\centering
\small
\caption{Two operating points at the deployed roster of four engines. Precision intervals are $95\%$ Wilson intervals, yield is correct labels divided by all input images, and the certified column gives simultaneous distribution-free lower bounds. The final row previews the frontier-model backstop of \S\ref{sec:backstop2}.}
\label{tab:gate}
\setlength{\tabcolsep}{3pt}
\begin{tabular}{lcccccc}
\toprule
gate
& accepted
& coverage
& correct
& precision
& yield
& certified $\geq$ \\
\midrule
\vlmstar{} alone
& $263$
& $100\%$
& $169$
& $0.643$ {\small$[.583,.698]$}
& $0.643$
& --- \\
$a\!\geq\!1$ (emit all)
& $263$
& $100\%$
& $196$
& $0.745$ {\small$[.689,.794]$}
& $0.745$
& $0.677$ \\
$a\!\geq\!2$ (\cons)
& $215$
& $81.7\%$
& $191$
& $0.888$ {\small$[.839,.924]$}
& $0.726$
& $0.829$ \\
$a\!\geq\!3$ (\gold)
& $137$
& $52.1\%$
& $135$
& $0.985$ {\small$[.948,.996]$}
& $0.513$
& $0.940$ \\
$a=4$ (unanimous)
& $59$
& $22.4\%$
& $58$
& $0.983$ {\small$[.910,.997]$}
& $0.221$
& $0.893$ \\
\midrule
$a\!\geq\!2$ $+$ frontier backstop
& $242$
& $92.0\%$
& $211$
& $0.872$ {\small$[.824,.908]$}
& $0.802$
& $0.814$ \\
\bottomrule
\end{tabular}
\end{table}

\FloatBarrier

%% ===========================================================================
%%
%%   4.1  fig:backstop + tab:backstop-rescue
%%   4.2  tab:standalone + tab:standalone-by-agreement
%%   5.1  fig:corpus + tab:corpus + tab:substantive-votes
%%   5.2  fig:recall
%%   5.3  tab:human-audit
%%   6.1  fig:kb + tab:kb
%%   6.2  tab:grounding
%% ===========================================================================

\section{Frontier-Model Backstopping}\label{sec:backstop}

%% ---------------------------------------------------------------------------
\subsection{Backstop utility is a property of the model version}
\label{sec:backstop2}

\begin{figure}[t]
\centering
\includegraphics[width=\textwidth]{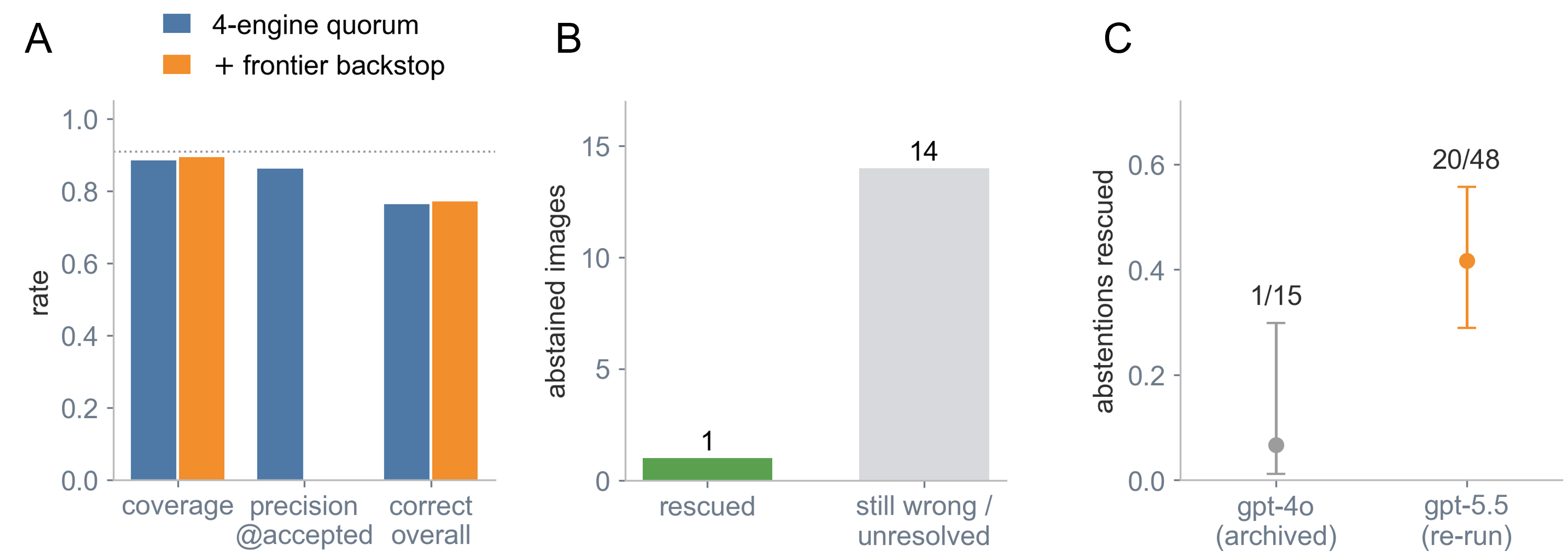}
\caption{The archived \texttt{gpt-4o} run resolves too few abstentions to be
interpreted, while the later \texttt{gpt-5.5} run resolves $20$ of $48$.}
\label{fig:backstop}
\end{figure}

Here, \texttt{gpt-4o} and then \texttt{gpt-5.5} were used on images that the quorum rejects. Figure~\ref{fig:backstop} shows both runs. The archived experiment used \texttt{gpt-4o} on an earlier $131$-image ACS subset, where the quorum accepted $116$ images at $0.862$ precision and abstained on $15$. The backstop resolved one of the $15$ and introduced no errors, raising coverage from $88.5\%$ to $89.3\%$ at a cost of \$$0.028$. A rescue rate of $1/15$ carries a Wilson interval of $[1.2\%,29.8\%]$, so the run determined nothing. The subset is also not a random sample of the main evaluation, its quorum coverage being $88.5\%$ against $81.7\%$, and the capping rule that produced it cannot be reconstructed. The experiment was therefore repeated at the scale of the main evaluation, using \texttt{gpt-5.5} on the $48$ images rejected at $a\!\geq\!2$. Predictions counted as correct only when the InChIKey matched the ground truth.

Table~\ref{tab:backstop-rescue} reports both runs. \texttt{gpt-5.5} resolves $20$ of $48$ abstentions, a rescue rate of $0.417$ $[.288,.557]$. Coverage rises from $81.7\%$ to $92.0\%$ and yield from $0.726$ to $0.802$, while precision falls from $0.888$ to $0.872$. Repeating the run three times per image gives the same $20$ rescues under majority voting, with $57$ of $144$ individual responses correct and answers varying on $11$ images. The two rows of Table~\ref{tab:backstop-rescue} are not a head-to-head comparison, since they use different subsets and their intervals overlap. What the pair establishes is that a backstop cannot be adopted or rejected on inherited evidence. Rescue rate is a joint property of the model version and the abstention set, and any deployment has to measure it on both.

\begin{table}[t]
\centering
\small
\caption{Rescue rates for the two backstop experiments. The runs use
different image subsets and are not a head-to-head comparison of the two
models.}
\label{tab:backstop-rescue}
\begin{tabular}{lccc}
\toprule
backstop experiment & abstentions & correct rescues & rescue rate ($95\%$ CI) \\
\midrule
\texttt{gpt-4o}, archived run
& $15$
& $1$
& $0.067$ {\small$[.012,.298]$} \\
\texttt{gpt-5.5}, current run
& $48$
& $20$
& $0.417$ {\small$[.288,.557]$} \\
\bottomrule
\end{tabular}
\end{table}

%% ---------------------------------------------------------------------------
\subsection{Agreement routes images to the system that can read them}
\label{sec:standalone}

Whether the frontier model can replace the consensus gate rather than backstop it is answered in Table~\ref{tab:standalone}: it cannot. The two-of-four quorum accepts more images ($81.7\%$ against $76.8\%$) at higher precision ($0.888$ against $0.812$) and higher yield ($0.726$ against $0.624$). With abstention disabled the quorum still leads, reaching $0.745$ against a yield of $0.624$. Adding the backstop raises coverage to $92.0\%$ at $0.872$ precision for an estimated \$$1.31$, against \$$5.13$ to run the frontier model alone.

Table~\ref{tab:standalone-by-agreement} shows why one system backstops the other instead of replacing it. The frontier model is better than the quorum by $29.2$ points where no two recognizers agree, and worse by $15$ to $26$ points everywhere else. The crossover is not incidental. An agreement count of one selects the images on which specialized recognizers fail, which is where a general model trained on a different distribution has an advantage. Once two recognizers agree, the same generality becomes a liability.

The agreement count therefore does more than accept and reject. It routes. Images with agreement are answered by the quorum, images without it are answered by a model that fails on different inputs, and the count identifies which case applies before either answer is trusted. This is the reason the backstop buys ten points of coverage for $1.6$ points of precision, and the reason substituting the frontier model for the quorum would lose precision across the bulk of the benchmark.

\begin{table}[t]
\centering
\small
\caption{The quorum exceeds the standalone frontier model on coverage,
precision, and yield, at a fraction of the cost. API costs assume
\$$2.5$/\$$10$ per million tokens and are approximate, since verified list
pricing is unavailable.}
\label{tab:standalone}
\setlength{\tabcolsep}{3.5pt}
\begin{tabular}{lccccc}
\toprule
system
& accepted
& coverage
& precision
& yield
& est.\ API cost \\
\midrule
\texttt{gpt-5.5} alone
& $202$
& $76.8\%$
& $0.812$ {\small$[.752,.860]$}
& $0.624$
& \$$5.13$ \\
\midrule
quorum, emit all
& $263$
& $100\%$
& $0.745$ {\small$[.689,.794]$}
& $0.745$
& --- \\
quorum $a\!\geq\!2$
& $215$
& $81.7\%$
& $0.888$ {\small$[.839,.924]$}
& $0.726$
& --- \\
quorum $a\!\geq\!3$
& $137$
& $52.1\%$
& $0.985$ {\small$[.948,.996]$}
& $0.513$
& --- \\
quorum $a\!\geq\!2$ $+$ backstop
& $242$
& $92.0\%$
& $0.872$ {\small$[.824,.908]$}
& $0.802$
& \$$1.31$ \\
\bottomrule
\end{tabular}
\end{table}

\begin{table}[t]
\centering
\small
\caption{The two systems fail on different images. The frontier model leads
only where no two recognizers agree. Refusals count as incorrect.}
\label{tab:standalone-by-agreement}
\setlength{\tabcolsep}{4pt}
\begin{tabular}{lcccc}
\toprule
system
& $a=1$ ($n=48$)
& $a=2$ ($n=78$)
& $a=3$ ($n=78$)
& $a=4$ ($n=59$) \\
\midrule
quorum correct
& $0.104$
& $0.718$
& $0.987$
& $0.983$ \\
\texttt{gpt-5.5} correct
& $0.396$
& $0.500$
& $0.731$
& $0.831$ \\
frontier $-$ quorum
& $+29.2$ pt
& $-21.8$ pt
& $-25.6$ pt
& $-15.3$ pt \\
\bottomrule
\end{tabular}
\end{table}

\FloatBarrier

%% ===========================================================================
\section{Journal-Domain Corpus}\label{sec:corpus}

%% ---------------------------------------------------------------------------
\subsection{The gate converts figures into labels by rejecting most of them}
\label{sec:pipeline}

\begin{figure}[t]
\centering
\includegraphics[width=\textwidth]{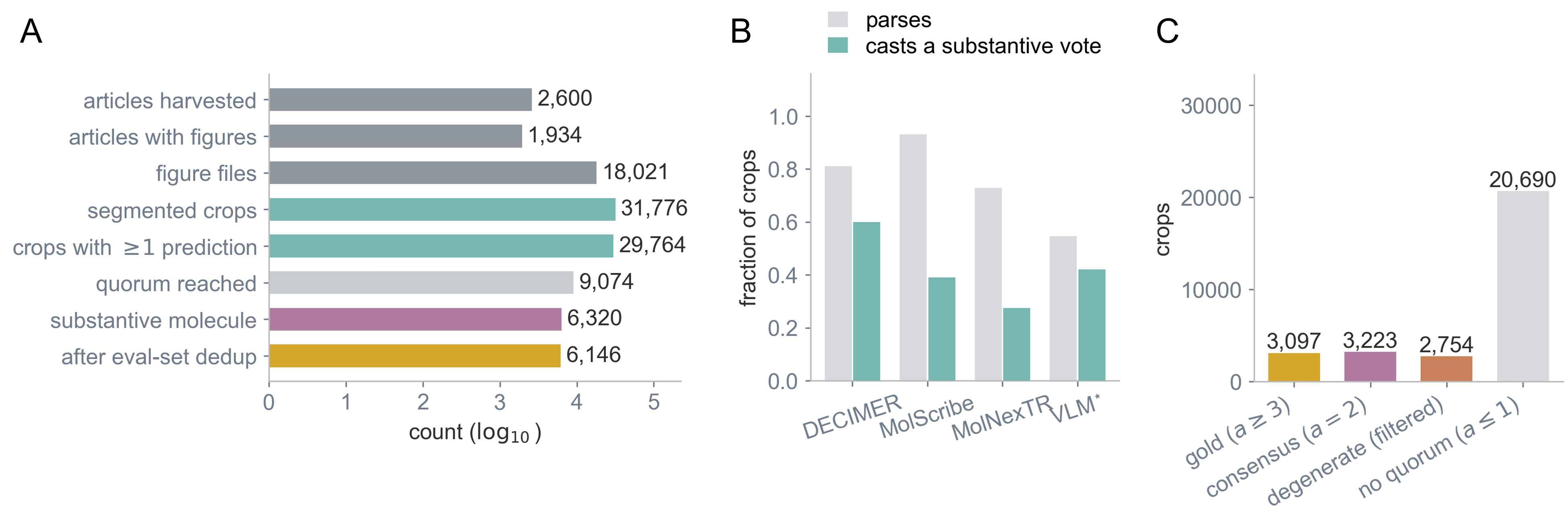}
\caption{The pipeline converts PMC figures into $6{,}146$ released structure
labels after consensus filtering and benchmark deduplication.}
\label{fig:corpus}
\end{figure}

Public OCSR training data are largely synthetic or patent-derived. VERDICT is validated for generating high confidence labels for molecular depictions after consensus filtering. Therefore, it is used to build a journal-domain corpus, $2{,}600$ chemistry-related PMC Open Access articles \citep{pmcoa} were collected, of which $1{,}934$ contain figures, giving $18{,}021$ figure files. DECIMER-Segmentation \citep{decimerseg} produced $31{,}776$ candidate crops, and $29{,}764$ received at least one parseable prediction. Figure~\ref{fig:corpus} traces what happens next.

Table~\ref{tab:substantive-votes} shows that parseability is a misleading measure of what an engine contributes. MolScribe returns a parseable string for $93.0\%$ of crops but a substantive molecular vote for $39.1\%$, while DECIMER returns parseable output less often and substantive output most often. The ranking by parseable output is close to the reverse of the ranking by substantive vote. Any pipeline that selects or weights engines by parse rate will therefore favor the engines that contribute least.

\begin{table}[t]
\centering
\small
\caption{Substance filtering removes all $2{,}193$ labels added by the
broader matching rule and returns both matching rules to the same corpus.}
\label{tab:corpus}
\setlength{\tabcolsep}{4.5pt}
\begin{tabular}{lccccc}
\toprule
stage
& \gold{} ($a\!\geq\!3$)
& \cons{} ($a=2$)
& labels
& distinct molecules
& kept \\
\midrule
strict InChIKey matching
& $3{,}215$
& $3{,}666$
& $6{,}881$
& ---
& $23.1\%$ \\
$+$ canonical-SMILES fallback
& $3{,}215$
& $5{,}859$
& $9{,}074$
& ---
& $30.5\%$ \\
\midrule
substance-filtered
& $3{,}097$
& $3{,}223$
& $6{,}320$
& $4{,}944$
& $21.2\%$ \\
$+$ evaluation-set deduplication
& $3{,}005$
& $3{,}141$
& $6{,}146$
& $4{,}833$
& $20.6\%$ \\
\bottomrule
\end{tabular}
\end{table}

\begin{table}[t]
\centering
\small
\caption{Parse rate inverts the ranking of engine contribution. The engine
that parses most often supplies the fewest substantive votes.}
\label{tab:substantive-votes}
\setlength{\tabcolsep}{5pt}
\begin{tabular}{lcccc}
\toprule
output measure & DECIMER & MolScribe & MolNexTR & \vlmstar{} \\
\midrule
parseable output
& $81.0\%$
& $93.0\%$
& $72.9\%$
& $54.5\%$ \\
valid identity key
& $81.0\%$
& $61.4\%$
& $38.9\%$
& $54.0\%$ \\
substantive vote
& $60.0\%$
& $39.1\%$
& $27.4\%$
& $42.1\%$ \\
\bottomrule
\end{tabular}
\end{table}

%% ---------------------------------------------------------------------------
\subsection{Segmentation sets the ceiling the gate cannot raise}
\label{sec:recall}

The pipeline is asymmetric in a way that determines how it should be tuned. An extra crop is harmless, because the gate rejects it. A structure that is never segmented is lost permanently, because no later stage can recover what was not cut out. Segmentation recall is therefore a ceiling on the corpus, while gate precision is only a floor on its quality.

\begin{figure}[t]
\centering
\includegraphics[width=0.55\textwidth]{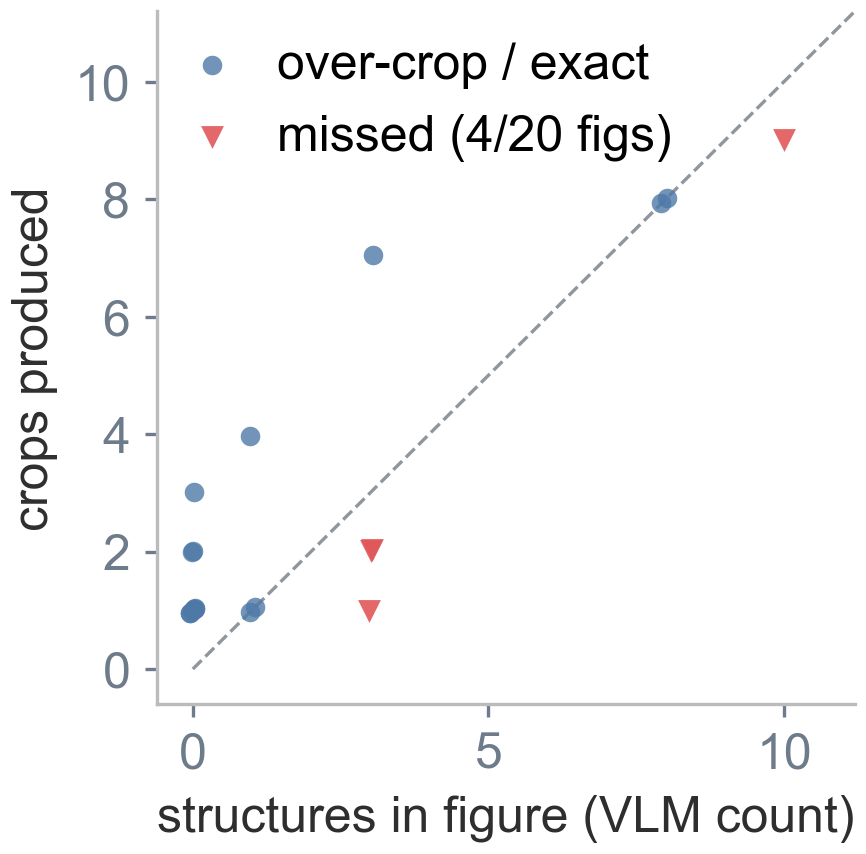}
\caption{Crops produced against the VLM reference count for each of $20$
probe figures. Points above the identity line are over-segmented and points
below have lost structures. Estimated recall is $0.878$, and most errors are
extra crops rather than omissions.}
\label{fig:recall}
\end{figure}

Recall was estimated on a $20$-figure probe in which a frontier VLM counted $41$ structures and DECIMER-Segmentation recovered $36$, giving $0.878$ ($95\%$ Wilson CI $[0.745,0.947]$). Figure~\ref{fig:recall} plots crops produced against the reference count for each figure. Sixteen figures lie on or above the identity line and four below it, so the dominant error is over-segmentation, which is the direction the pipeline tolerates. Several figures with no reference structure still produced one to three crops, all of which the gate later removed. The densest figure, with ten structures, lost one. The estimate rests on a small sample and on a VLM reference count rather than human annotation, so it bounds the plausible recall rather than measuring it.

\FloatBarrier

%% ---------------------------------------------------------------------------
\subsection{Human review confirms the tiers and finds what they miss}
\label{sec:audit}

Precision measured on ACS need not transfer to PMC, where no ground truth exists. Two non-overlapping samples of $200$ released labels were drawn, each containing $100$ \gold{} and $100$ \cons{} labels, and a chemist blinded to the tier marked every crop as correct, wrong, or unscoreable.

Table~\ref{tab:human-audit} reports the result. The two samples agree closely. Pooled precision is $0.995$ for \gold{} and $0.958$ for \cons{}, and unscoreable labels are nine times more common in \cons{}. The \gold{} figure matches its ACS estimate, while \cons{} performs $24$ percentage points better on PMC than on ACS. The thresholds are therefore conservative outside the domain that produced them rather than fitted to it, which is the direction an operating point should err.

The audit also found a defect the substance filter had missed. Outputs such as \texttt{O.O.O.O}\ldots{} parse cleanly, contain no dummy atoms, and clear the heavy-atom threshold by repetition, without representing one molecule. A scan of the corpus found $20$ such labels among $6{,}320$ ($0.32\%$). All lie in \cons{} and none in \gold{}. The corrected rule rejects any label containing three or more copies of the same disconnected component, and leaves \gold{} unchanged. A defect discovered independently of the tier system respecting that system is stronger evidence for the tier ordering than the precision figures alone.

\begin{table}[t]
\centering
\small
\caption{Human review measures $99.5\%$ precision for \gold{} and $95.8\%$ for \cons{} across two independent samples. Intervals are $95\%$ Wilson intervals over scoreable labels.}
\label{tab:human-audit}
\setlength{\tabcolsep}{4pt}
\begin{tabular}{llccccc}
\toprule
tier
& sample
& correct
& wrong
& unscoreable
& precision
& $95\%$ CI \\
\midrule
\gold{}
& first
& $99$
& $1$
& $0$
& $0.990$
& $[.946,.998]$ \\
& second
& $99$
& $0$
& $1$
& $1.000$
& $[.963,1.00]$ \\
& \textbf{pooled}
& $\mathbf{198}$
& $\mathbf{1}$
& $\mathbf{1}$
& $\mathbf{0.995}$
& $\mathbf{[.972,.999]}$ \\
\midrule
\cons{}
& first
& $91$
& $4$
& $5$
& $0.958$
& $[.897,.984]$ \\
& second
& $92$
& $4$
& $4$
& $0.958$
& $[.898,.984]$ \\
& \textbf{pooled}
& $\mathbf{183}$
& $\mathbf{8}$
& $\mathbf{9}$
& $\mathbf{0.958}$
& $\mathbf{[.920,.979]}$ \\
\bottomrule
\end{tabular}
\end{table}

\FloatBarrier

%% ===========================================================================
\section{A Grounded Molecular Knowledge Base}\label{sec:kb}

%% ---------------------------------------------------------------------------
\subsection{Label quality propagates into downstream coverage}
\label{sec:kbcoverage}

Each accepted molecule becomes a record holding its SMILES, structure crop, source article, PMCID, and figure number. A frontier model then extracts the compound name, physical properties, synthesis information, bioactivity, a summary, and supporting text from the article. This is the only stage in the system that requires a paid model call.

\begin{figure}[t]
\centering
\includegraphics[width=0.86\textwidth]{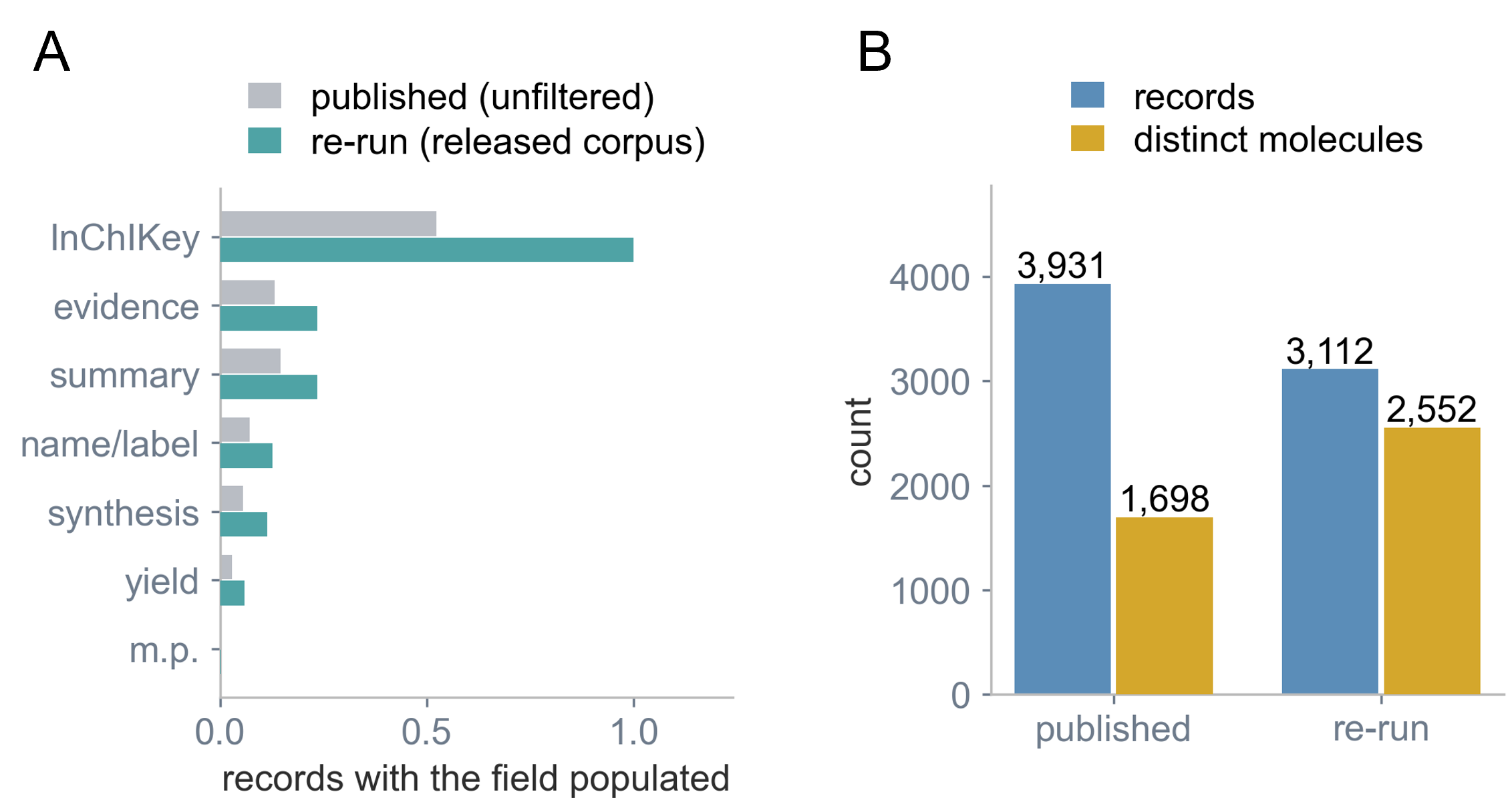}
\caption{Filtered structure labels increase distinct-molecule coverage and roughly double the population of most knowledge-base fields.}
\label{fig:kb}
\end{figure}

The effect of label quality was isolated by running the extraction twice. The first run used an older set of $10{,}896$ labels produced before the four-engine merge and the substance filter. The second used the released $6{,}146$. Both runs share the same $100$ articles, prompt, model, and limit of $50$ molecules per article, so the labels are the only material difference. Figure~\ref{fig:kb} and Table~\ref{tab:kb} report the outcome.

Fewer labels produce fewer records but more chemistry. Records fall from $3{,}931$ to $3{,}112$ while distinct molecules rise from $1{,}698$ to $2{,}552$, because invalid identity keys had been collapsing unrelated crops onto shared records. Summary coverage rises from $14.7\%$ to $23.5\%$, synthesis coverage from $5.6\%$ to $11.4\%$, and evidence coverage from $13.3\%$ to $23.6\%$, at an unchanged cost of about \$$7.80$ per $100$ articles. Field sparsity that would ordinarily be attributed to the source literature was therefore caused in part by the labels. Errors upstream of an extraction step do not merely pass through it. They consume its budget.

\begin{table}[t]
\centering
\small
\caption{Filtering the input labels increases molecule and field coverage at unchanged extraction cost.}
\label{tab:kb}
\setlength{\tabcolsep}{5pt}
\begin{tabular}{lcc}
\toprule
& original run & corrected run \\
\midrule
input labels
& $10{,}896$ (unfiltered)
& $6{,}146$ (filtered) \\
records
& $3{,}931$
& $3{,}112$ \\
distinct molecules
& $1{,}698$
& $2{,}552$ \\
\midrule
InChIKey
& $52.4\%$
& $100\%$ \\
evidence snippet
& $13.3\%$
& $23.6\%$ \\
text summary
& $14.7\%$
& $23.5\%$ \\
compound name
& $7.2\%$
& $12.6\%$ \\
synthesis route
& $5.6\%$
& $11.4\%$ \\
yield
& $3.0\%$
& $6.0\%$ \\
bioactivity
& $1.7\%$
& $3.3\%$ \\
melting point
& $0.3\%$
& $0.3\%$ \\
\midrule
numeric values extracted
& $150$
& $235$ \\
\quad verbatim in article
& $150$
& $216$ \\
\quad found unsupported
& $0$
& $0$ \\
\midrule
cost
& \$$7.958$ (\$$0.0796$/article)
& \$$7.804$ (\$$0.0780$/article) \\
\bottomrule
\end{tabular}
\end{table}

%% ---------------------------------------------------------------------------
\subsection{Traceability depends on the field type}
\label{sec:grounding}

Every extracted field was checked against the full text of its source article. Numeric values were tested for verbatim occurrence. Prose snippets were scored by the fraction of their $8$-grams appearing in the article and counted as well supported above $80\%$.

Table~\ref{tab:grounding} shows that traceability is not uniform across the schema. Of $235$ numeric fields, $216$ appear verbatim and a further $15$ appear in the article but under the wrong schema key, giving $98.3\%$ traceable and no unsupported number anywhere in the corpus. Compound names reach $81.4\%$, with most of the remainder being normalized forms or expanded abbreviations. Prose is weaker. Only $325$ of $653$ evidence snippets meet the threshold and $115$ have no verbatim overlap at all, a pattern that also holds in a matched comparison of $132$ molecule and article pairs. The cause of the lower prose overlap is not established.

The knowledge base is therefore typed by verifiability rather than treated uniformly. Numeric fields are released as traceable data, prose snippets as retrieval aids, and every snippet retains a link to the article it came from so that a reader can check it directly. A field that cannot be verified is still useful for finding the right paper, provided it is not presented as having been read from one.

\begin{table}[t]
\centering
\small
\caption{Numeric fields are traceable and compound names mostly are, while half of the prose snippets lack strong verbatim support.}
\label{tab:grounding}
\begin{tabular}{lrrl}
\toprule
field & evaluated & matched criterion & result \\
\midrule
numeric values
& $235$
& $231$ traceable
& $98.3\%$ \\
compound names
& $393$
& $320$ verbatim
& $81.4\%$ \\
evidence snippets
& $653$
& $325$ well supported
& $49.8\%$ \\
evidence snippets
& $653$
& $115$ with zero overlap
& $17.6\%$ \\
\bottomrule
\end{tabular}
\end{table}

\FloatBarrier

\section{Limitations}\label{sec:limits}

The comparison is model dependent. Frontier-model behavior changed substantially between the \texttt{gpt-4o} and \texttt{gpt-5.5} backstop runs, so the backstop and standalone results describe a model version rather than a capability, and both should be re-measured as frontier models are developing. The negative round-trip result is likewise bounded by the implementation of Appendix~\ref{app:rt}. Two controls would be required before it could be stated more broadly than a failure of this family of pixel-space comparison. The first is a rendering control: scoring a rendering of the ground-truth structure against the same crop, which separates failure of the comparison from failure of the predictions. The second is a registration control: repeating the comparison after scale, rotation, and stroke-width normalization, which separates the comparison metric from the alignment it assumes. Neither was run. Render-then-recognize, which is informative but weaker than agreement, is not covered by this limitation. The operating curve is therefore evidence for this engine mix and this evaluation period, not a permanent ranking of recognizers.

Validation remains limited in scope. The signal comparison uses $263$ ACS images, the backstop uses $48$ abstentions, and the segmentation probe uses $20$ figures with VLM-derived reference counts rather than human annotation. The cross-dataset experiments in \S\ref{sec:crossset} use the first $500$ archived rows of each set rather than random samples, so those coverage and precision values are not sampling estimates for the full benchmarks. Markush structures fall outside identity-key evaluation altogether, and the knowledge-base records have not received full chemical adjudication. Larger randomly sampled audits are needed to characterize rare errors and to support patent-domain deployment.

The provenance audit is molecule-level rather than image-level, because the original image manifests were not preserved. No ACS molecule overlaps \vlmstar{}'s training pool, but $15.0\%$ of corpus molecules do; the dependence is removable through external-only agreement or the public three-engine subset. Source licenses also differ across articles. Commercial users must filter on the recorded license field, and the corpus should not be treated as uniformly licensed for commercial use.

\section{Conclusion}
Deployable OCSR requires not only an accurate recognizer, but also a system that can determine which of its own predictions to keep. Within this system, agreement on molecule identity among architecturally distinct recognizers carries reliable information (AUROC $0.916$) measured on real labeled journal figures. VERDICT system converts agreement into a gate reaching $0.888$ precision at $81.7\%$ coverage and $0.985$ at $52.1\%$, with distribution-free lower bounds of $0.829$ and $0.940$, at $3.5$\,s per image and no marginal API cost. Its reliability does not rest on a single engine but rather depends on the independence of the various engines and the rules governing their coordination. For example, a loose matching rule once resulted in $2{,}193$ erroneous matches, all of which involve wildcards or R-group fragments rather than complete molecules. This system addressed such errors by molecule identity level matching, combined with explicit substance filtering. A portion of what the gate abstains on was recovered via a current frontier model \texttt{gpt-5.5}, raising coverage to $92.0\%$ at $0.802$ yield. VERDICT further performed on PMC journals and outputted $6{,}146$ deduplicated structure labels, where a chemist evaluated $400$ of them reporting at $0.995$ precision for \gold{} and $0.958$ for \cons{}, and $94.7\%$ of them reproducible from public components alone. After a full evaluation, a corpus obtained this way ceases to be a weak-label pool and becomes a training set of real labeled depictions for a model with higher accuracy itself \citep{ft}. Because every label retains the publication it came from, the same corpus sustains a cited, queryable knowledge base, and a structure that existed only as ink in a single figure becomes reachable alongside the text and measurements recorded around it.

\bibliographystyle{plainnat}
\bibliography{references}

\appendix

\section{The reconciliation core}\label{app:engine}
The core is deterministic given fixed engine outputs and imports no model SDK; engines are
adapters implementing \texttt{predict(image)\,$\rightarrow$\,SMILES\,$|$\,None} and are
selected at runtime, so adding an engine never touches the voting logic. This matters
practically: DECIMER requires TensorFlow, MolScribe and MolNexTR require different PyTorch
builds, and \vlmstar{} requires a serving stack, so the four cannot share a Python
environment. Adapters run in separate environments behind a uniform interface and the core
consumes their outputs.

Reconciliation is the InChIKey mode of \S\ref{sec:verdict} with deterministic tie-breaking
(highest agreement, then a fixed engine priority) and the substance check of
\S\ref{sec:keybug}, so a given set of engine outputs always produces the same label and tier.
Latency on the ACS run, four engines in parallel on one node: median $3.49$\,s, $p90$ $6.92$\,s,
$p95$ $9.27$\,s, mean $4.68$\,s.

\paragraph{Two senses of reproducible, kept separate here.} The reconciliation core is
deterministic given engine outputs, so every number in this paper recomputes exactly
from the archived prediction files---that is the guarantee the Reproducibility note makes, and
App.~\ref{app:repro} names the scripts. Whether re-running the engines reproduces those
predictions is a separate question. All four decode greedily---\vlmstar{} generates with
\texttt{do\_sample=False} and $256$ max new tokens, MolScribe and MolNexTR use their default
greedy decoders, DECIMER its default beam---so there is no sampling temperature to fix, and
residual variation would come from library and driver versions rather than from the models.
The archived runs record checkpoint paths but not a full dependency lock, which is one reason
the re-run discussed in \S\ref{sec:limits} is a reconstruction rather than a replay.

That residual has since been measured directly. The three open engines were run over the same
$331$ images twice, once on CPU and once on GPU, in separately built environments. Of the $993$
resulting predictions, $988$ are character-identical. Four of the five differences are
alternative spellings of output that fails to parse under either run, and the fifth is one
MolScribe image where one run emits an unparseable string and the other a parseable but
incorrect one; that image is scored wrong either way. MolScribe's confidence scores agree to
$3\times10^{-6}$. Every per-engine accuracy quoted in this paper is therefore identical
across the two runs at integer-count resolution---$99$, $186$ and $162$ correct of $263$---and
the confidence AUROC agrees to four decimals. Engine-level replay is reproducible in the sense
that matters for the numbers reported here, though not character-for-character.

\section{The round-trip score no longer gated on}\label{app:rt}
For completeness, the signal measured in \S\ref{sec:signals}. A candidate SMILES is rendered
deterministically with RDKit \citep{rdkit}; both the input crop and the render are converted to
grayscale, resized to $256\times256$, and compared by an equally weighted mean of global SSIM
\citep{ssim} and the Jaccard overlap of ink masks binarized at intensity $200$:
\[
s_{\mathrm{rt}}=\tfrac12\,\mathrm{SSIM}(I,R)+\tfrac12\,\mathrm{IoU}\!\left(1[I<200],\ 1[R<200]\right).
\]
It is CPU-only, deterministic, and learning-free---all the properties that make it appealing.
The earlier deployed gate fused it with agreement as
$c=\sigma(4a+4s_{\mathrm{rt}}-4)$, accepting at $c\geq0.60$. Since $s_{\mathrm{rt}}$ occupies
$[0.0005,0.098]$ on real crops (\S\ref{sec:signals}), its contribution to the logit is at most
$0.39$ against agreement's $4$ per vote: on real documents the fused gate was already an
agreement gate with noise added, which is now stated explicitly.

\paragraph{What this implementation does not do.} It compares at a fixed $256\times256$ with no
registration, no scale or rotation search, no matching of bond length, line width or font, and
a single global SSIM rather than a local or masked one. Each of those is a plausible reason a
better implementation could separate correct from incorrect predictions where this one does
not, and none were tested; \S\ref{sec:limits} lists the two controls considered necessary
before the negative result can be stated more broadly than ``this family of pixel-space
comparison fails on real depictions''. What can be said beyond the AUROC is that the failure is
not a threshold choice: the oracle-tuned cut point in \S\ref{sec:signals} is dominated by the
trivial emit-everything baseline, so the score is not a good signal being read at the wrong
operating point.

\section{Corpus datasheet}\label{app:corpus}
Source. PMC Open Access subset \citep{pmcoa}; chemistry-dense journals; article full
text via E-utilities JATS, figures and SI via the OA package. Scale. $2{,}600$
articles harvested, $1{,}934$ with figures, $18{,}021$ figure files, $31{,}776$ segmented
crops, $29{,}764$ with at least one parseable prediction. Labels. $9{,}074$ reach a
quorum, $6{,}320$ survive the substance filter (\gold{} $3{,}097$ / \cons{} $3{,}223$), and
$6{,}146$ survive evaluation-set deduplication, over $4{,}833$ distinct
molecules from $476$ articles. Per-record fields. crop path,
consensus SMILES, InChIKey, tier, agreement count $a$, external agreement
$a_{\mathrm{ext}}$, per-engine raw predictions, PMCID and figure id. Known biases.
RSC/ACS-dominated venue mix; over-representation of small aromatic intermediates; systematic
absence of Markush and reaction-scheme content, which the gate rejects. Intended
use. training and analysis; the \gold{} slice is additionally usable for evaluation, its
precision having been measured at $0.995$ $[.972,.999]$ on $400$ adjudicated labels across two
independent draws (\S\ref{sec:corpus}).
Licensing. The $476$ source articles carry four terms---CC-BY-4.0, CC-BY-3.0, CC-BY-NC-3.0
and CC-BY-NC-ND-4.0---so the collection has no single license; $5{,}598$ of $6{,}320$ labels
($88.6\%$) permit commercial use. Derived labels are redistributed
(SMILES, InChIKey, tier, $a$, $a_{\mathrm{ext}}$, per-engine raw predictions, PMCID, figure id)
together with the per-article license identifier and a commercial-use flag, and not the images;
every crop is reconstructible from the PMC OA package using the shipped PMCID and figure id plus
the segmentation step of \S\ref{sec:corpus}. Commercial users must filter on the license field.
Reproducing without the in-house model. A three-engine variant built from public components
only (DECIMER, MolScribe, MolNexTR) yields $5{,}984$ labels, $94.7\%$ of the released total;
the released files carry $a_{\mathrm{ext}}$ so this subset can be selected without re-running
anything.

\section{Reproducing every number}\label{app:repro}

\paragraph{Availability.} The corpus, the reconciliation core with its per-engine adapters, and
the analysis scripts below are deposited in a public archive under a persistent DOI.
\ifdepositlive
Archive: \texttt{https://doi.org/\zenododoi}, released under CC-BY-4.0.
  \ifrepolive
  The code is also mirrored on GitHub at \texttt{\repourl}.
  \else
  Every script named below is contained in that archive.
  \fi
\else
The identifier is stated in the Data and Code Availability statement of the published version
and is omitted here only because this manuscript is under review.
\fi
Nothing in the analysis depends on cluster access: every file named below ships with
the paper source, with the one exception noted at the end of this appendix. The deposit
redistributes derived labels and scripts, never images; the manuscript figures are likewise not
included, because their final panels were laid out by hand from the per-panel outputs of
\texttt{make\_verdict\_figs.py}, so the script rather than the assembled artwork is the
reproducible object.

The artifacts, copied from the cluster PVC into \texttt{data/} in the paper source, are:

\begin{itemize}\itemsep1pt
\item \texttt{acs\_consensus\_full.csv} --- the $n=263$ live four-engine run, one row per image:
ground-truth key, agreement count, winning key and correctness, round-trip score, \vlmstar{}
prediction, latency. Everything in \S\ref{sec:signals} comes from this file.
\item \texttt{weak\_labels\_4eng\_identitykey.csv} --- the $9{,}074$ quorum-reaching corpus
labels with tier, before the substance filter, so the audit of \S\ref{sec:keybug} is
reproducible rather than merely asserted.
\item \texttt{rag\_records.csv} --- the $3{,}931$ knowledge-base records of the published
extraction pass, and \texttt{rag\_records\_filtered.csv} --- the $3{,}112$ records of the
corrected re-run over the released label set (\S\ref{sec:kb}, Table~\ref{tab:kb}).
\item \texttt{backstop\_current.json} --- the per-image results of the \texttt{gpt-5.5} run on
the $48$ abstentions (\S\ref{sec:backstop2}), including its raw prediction for each.
\item \texttt{contamination.json} --- the molecule-level training-pool audit
(\S\ref{par:contamination}).
\item \texttt{openai\_missrate.json} --- the $20$-figure segmentation-recall probe.
\item \texttt{robustness.log} --- the frontier-backstop ablation report (\S\ref{sec:backstop}).
\item \texttt{detail\_4eng.csv} --- the per-engine predictions for all $29{,}764$ crops, which
is what makes the per-engine analysis of \S\ref{sec:corpus}, the pairwise-agreement measurement
of \S\ref{sec:signals} and the three-engine ablation reproducible without a GPU.
\item \texttt{corpus\_counts.json} --- funnel counts measured on the PVC, each with the command
that produced it.
\end{itemize}

\texttt{rag\_extract\_filtered.py} re-runs the
knowledge-base extraction over the released label set (\S\ref{sec:kb}) and
\texttt{kb\_compare.py} produces Table~\ref{tab:kb} and the grounding numbers;
\texttt{backstop\_current.py} runs \S\ref{sec:backstop2}; \texttt{contamination\_effect.py} and
\texttt{build\_release\_corpus.py} produce the exposure analysis and the shipped corpus file;
\texttt{revision3\_stats.py} derives every quantity quoted in the revised \S\ref{sec:backstop2}
and \S\ref{sec:kb} from those raw outputs. Of the earlier scripts,
\texttt{make\_verdict\_figs.py} recomputes the quantities of
\S\ref{sec:signals}--\ref{sec:kb}. \texttt{revision\_stats.py} covers the AUROC intervals, the oracle-threshold
probe, the Learn-then-Test bounds, the degenerate-set bifurcation and threshold sweep,
per-engine behaviour and the engine ablation. \texttt{revision2\_stats.py} covers the
article-level cluster bootstrap, the unconditioned pairwise agreements, the substantive-vote
rates and the $a\!=\!2$ pair composition. \texttt{contamination\_check.py} performs the
molecule-level training-pool audit of \S\ref{par:contamination} and the scaffold overlap of
\S\ref{sec:corpus}. \texttt{dedup.py} performs the evaluation-set deduplication. The
per-engine re-run has its own chain: \texttt{merge\_acs\_per\_engine.py} assembles the
prediction table, \texttt{per\_engine\_analysis.py} derives the single-engine accuracies, the
intrinsic-confidence AUROC of \S\ref{sec:signals}, the error-correlation and three-engine
quorum figures and the reproduction check against the archived gate, and
\texttt{rerun\_reconcile.py} produces the Markush false-accept counts with and without the
substance filter, the Clopper--Pearson bound on them, and the CPU-versus-GPU replay comparison
reported above. The render-then-recognise control of \S\ref{sec:signals} adds two more:
\texttt{render\_predictions.py} draws each engine's prediction and each ground-truth structure
under a fixed RDKit configuration, and \texttt{analyze\_rtr.py} scores self-consistency, its
AUROC and bootstrap interval, the accept-if-consistent operating point and the gold-render
controls. \texttt{crossset\_gate.py} recomputes the operating curve of \S\ref{sec:crossset} on
all four benchmarks from the archived per-engine predictions. Wilson
intervals \citep{wilson}, Clopper--Pearson bounds \citep{clopperpearson} and DeLong intervals
\citep{delong} are computed in these scripts; Mann--Whitney and Fisher tests use SciPy. Any
discrepancy between a table and these scripts' output is a bug in the table.

Two measurements need files too large to ship with the manuscript and are run against the
archive instead: the contamination audit reads the $365{,}776$-row training-pool manifest, and
the extraction-grounding check of \S\ref{sec:kb} reads the $100$ source articles' full text
($\approx\!1.4$\,GB). Both scripts, their exact invocations and their outputs are included in
the deposit; their results are transcribed into \texttt{data/contamination.json} and
\texttt{data/revision\_stats.json} with that provenance recorded, and they are flagged explicitly
rather than implying the whole analysis runs from the shipped \texttt{data/} directory.

\end{document}